\documentclass[letterpaper, 10 pt, conference]{ieeeconf}  % Comment this line out if you need a4paper

\IEEEoverridecommandlockouts                              % This command is only needed if 
\usepackage[english]{babel}
\usepackage[utf8]{inputenc}

\usepackage{graphicx}
\usepackage{subcaption} % for pdf, bitmapped graphics files
\usepackage{epsfig} % for postscript graphics files
\usepackage{times} % assumes new font selection scheme installed
\usepackage{svg}

\usepackage{url}
\usepackage{float}

\usepackage{amsmath, amsfonts, amssymb, amsthm}

\usepackage{algorithm}
\usepackage{algorithmic}

\usepackage[hidelinks]{hyperref}

\usepackage[colorinlistoftodos]{todonotes}

\newcommand{\etal}{\textit{et al.}}

\newcommand{\ie}{\textit{i.e.}}

\DeclareMathOperator{\atantwo}{atan2}

\newcommand{\peopleset}{\mathcal{N}}
\newcommand{\numpersons}{h}

\newcommand{\individual}[1]{\ensuremath{\mathbf{n}_{#1}}}

\newcommand{\pos}{\ensuremath{\mathbf{p}}}
\newcommand{\config}[1]{\ensuremath{\mathbf{q}_{#1}}}

\newcommand{\mmconf}[1]{\ensuremath{\mathbf{q}^\textrm{\tiny MM}_{#1}}}
\newcommand{\basepos}[1]{\ensuremath{\mathbf{p}^\textrm{B}_{#1}}}

\title{\LARGE \bf
Socially-aware Object Transportation by a Mobile Manipulator\\in Static Planar Environments with Obstacles
}

\author{Caio C. G. Ribeiro \qquad Leonardo R. D. Paes \qquad Douglas G. Macharet  % <-this % stops a space
\thanks{This work was supported by CAPES/Brazil - Finance Code 001, CNPq/Brazil, and FAPEMIG/Brazil.}
\thanks{The authors are with the Computer Vision and Robotics Laboratory (VeRLab), Department of Computer Science, Universidade Federal de Minas Gerais, MG, Brazil. E-mails: {\tt\small caioconti@ufmg.br, leonardordpaes@ufmg.br, doug@dcc.ufmg.br}
}}

\begin{document}
% IEEE Copyright Notice---
\onecolumn
    {\Large \textbf{IEEE Copyright Notice}}\\[2em]
    
    © 2025 IEEE. Personal use of this material is permitted. Permission from IEEE must be obtained for all other uses, in any current or future media, including reprinting/republishing this material for advertising or promotional purposes, creating new collective works, for resale or redistribution to servers or lists, or reuse of any copyrighted component of this work in other works.\\[1.5em]
    
    \textit{Accepted to be published in: 2025 34th IEEE International Conference on Robot and Human Interactive Communication (RO-MAN).}
\twocolumn
%------

\maketitle
\thispagestyle{empty}
\pagestyle{empty}

%%%%%%%%%%%%%%%%%%%%%%%%%%%%%%%%%%%%%%%%%%%%%%%%%%%%%%%%%%%%%%%%%%%%%%%%%%%%%%%%

\begin{abstract}

Socially-aware robotic navigation is essential in environments where humans and robots coexist, ensuring both safety and comfort. However, most existing approaches have been primarily developed for mobile robots, leaving a significant gap in research that addresses the unique challenges posed by mobile manipulators.
In this paper, we tackle the challenge of navigating a robotic mobile manipulator, carrying a non-negligible load, within a static human-populated environment while adhering to social norms. Our goal is to develop a method that enables the robot to simultaneously manipulate an object and navigate between locations in a socially-aware manner.
We propose an approach based on the Risk-RRT* framework that enables the coordinated actuation of both the mobile base and manipulator. This approach ensures collision-free navigation while adhering to human social preferences.
We compared our approach in a simulated environment to socially-aware mobile-only methods applied to a mobile manipulator. The results highlight the necessity for mobile manipulator-specific techniques, with our method outperforming mobile-only approaches. Our method enabled the robot to navigate, transport an object, avoid collisions, and minimize social discomfort effectively.

%We evaluate our technique in a simulated environment against the standard RRT* and Risk-RRT* methods that focus solely on the robot's base. The experiments demonstrate that our approach effectively enables the robot to navigate and transport an object to the desired location while avoiding collisions and respecting social distances whenever possible, or minimizing social discomfort when it is not.

\end{abstract}

%%%%%%%%%%%%%%%%%%%%%%%%%%%%%%%%%%%%%%%%%%%%%%%%%%%%%%%%%%%%%%%%%%%%%%%%%%%%%%%%

\section{Introduction}
        The presence of robots in human-shared environments has been steadily increasing over the past few years and is expected to continue growing \cite{Mavrogiannis2023,Silva2023}. Robots are increasingly cooperating with people on joint tasks or simply coexisting in shared spaces, such as factories~\cite{Thakar2018} and hospitals~\cite{Mahdi2022}.
        %, and schools~\cite{Ahtinen2020}.
        
        In these settings, robots must navigate not only to avoid humans and obstacles but also to ensure human comfort~\cite{Chi2018,Hirose2023}. This consideration is crucial for promoting safety and fostering acceptance of autonomous vehicles in everyday environments. Consequently, developing techniques for socially-aware navigation represents a significant advancement in making interactions between robots and people more harmonious and effective.

        %For such tasks, it is essential to quantify social acceptance to be used as a metric. Due to the subjective nature of humans, there are not fully-agreed answers in this problem. It is even possible to find works with opposite approaches, while most works agree that the front of a person is more inconvenient for a robot to be \cite{Kirby2010}, the paper by \cite{Svenstrup2010} argues that the best place to navigate is in the front due to visibility. However, the concept of personal space (\emph{Proxemics}) by Hall \cite{Hall1966} is widely referenced, with further works \cite{Kirby2010} improving its approach, which will be used in this work.

        %Relying on Proxemics or on different metrics, various approaches have been developed for social navigation (\cite{AlineSilva2023},\cite{Chi2018},\cite{Hirose2023}), employing techniques such as Social Forces \cite{Patompak2016}, Reinforcement Learning \cite{Dong2024}, and Sampling-based methods \cite{Silva2023}. However, the vast majority of found works focus solely on mobile robots. The domain of social navigation for mobile manipulators remain under explored in comparison.

\begin{figure}[t]
            \centering
            \begin{subfigure}{0.47\linewidth} % Adjust the width to fit in a single column
                \centering
                \includegraphics[width=\linewidth, trim={0.75cm 0cm 1.25cm 0.1cm}]{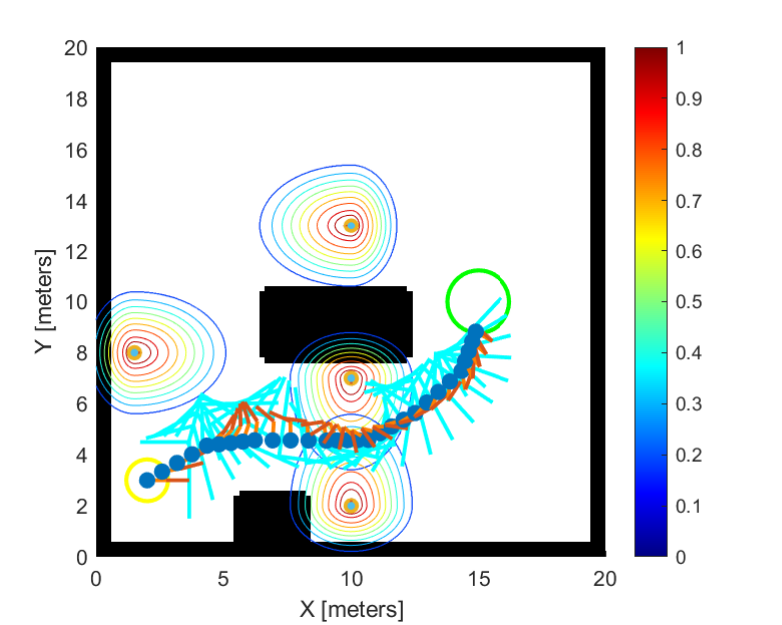} % Replace with your figure file
                \caption{Non-social Navigation}
                \label{fig:subfigure1}
            \end{subfigure}
            \hspace{1mm}
            %\hfill
            \begin{subfigure}{0.47\linewidth} % Adjust the width to fit in a single column
                \centering
                \includegraphics[width=\linewidth, trim={0.75cm 0cm 1.25cm 0.1cm}]{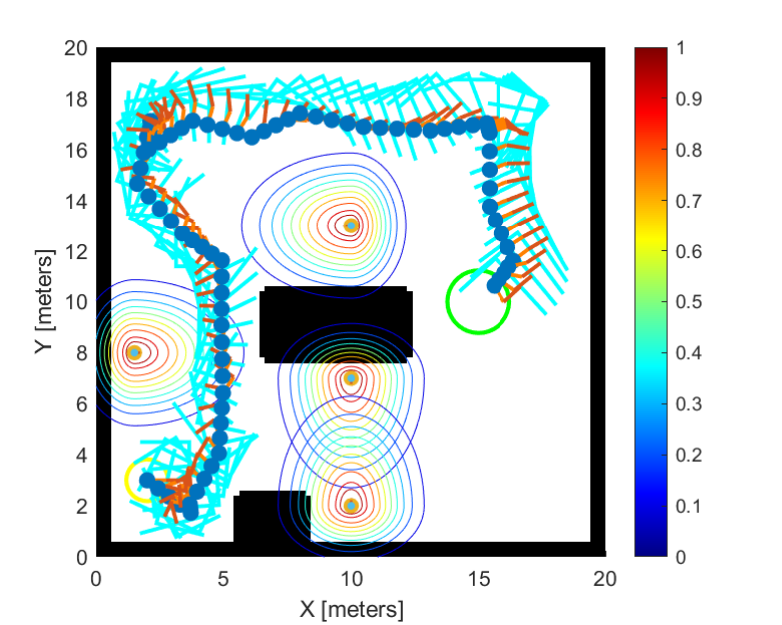} % Replace with your figure file
                \caption{Social Navigation (ours)}
                \label{fig:subfigure2}
            \end{subfigure}
            \caption{
            Illustrative scenario with four individuals dispersed in a cluttered environment, where their social regions are visualized. The color bar indicates the social cost around each person. The robot starts with its base at the center of the yellow circle and must reach the green circle. The robot is the dark blue circle with two links in orange, carrying the light-blue object.
            }
            \label{fig:initial_figure}
        \end{figure}

        Mobile manipulators, which integrate mobility with robotic manipulation, represent a significant leap in robotics, offering enhanced versatility for tasks in diverse environments. By combining a mobile base with a manipulator arm, these robots can perform complex tasks such as transporting materials in warehouses or assisting in delicate procedures in healthcare settings. As their use expands, ensuring socially-acceptable operation becomes increasingly crucial. This involves not only managing the base motion but also considering the robotic arm and the object being transported. Advancing planning and control techniques for these robots is essential to maximize their effectiveness and foster broader acceptance in human-centric environments.
        Figure~\ref{fig:initial_figure} presents an example that highlights the importance of specialized methods to achieve socially-acceptable object transportation.
        
        %To address socially-aware mobile manipulators, especially if the robot is carrying a dangerous load, additional concerns arise. It is crucial to consider not just the mobile base but also the robotic arm and the object being transported for a complete social navigation. This introduces the problem to generate a social whole-body motion planner that will account for the base, arm and object. In this work, we tackle this problem.

        %Addressing socially-aware mobile manipulation, particularly when the robot is carrying a potentially hazardous load, introduces additional complexities. It is essential to account for not only the mobile base but also the robotic arm and the object being transported to ensure comprehensive social navigation. This requires developing a social motion planner that integrates the movements of the base, arm, and object. 

        In this work, we tackle this challenge by proposing a framework built around a socially-aware whole-body motion planner that accounts for the mobile base, attached manipulator, and transported object. %\rev{We use our method in a static human-populated environment. Although we consider static humans, which is a simplification, office-like environments tend to have predictable human presence around desks and chairs. Thus, our method can pre-plan socially-aware paths based on expected human locations, while dynamic adaptations can be handled by complementary techniques using our approach as a foundation. Despite this simplification, our work represents an important step toward developing social navigation strategies for mobile manipulators, particularly when transporting large objects that may cause discomfort.} 
        We apply our method in a static human-populated environment. While assuming static humans is a simplification, office-like settings often have predictable human presence around desks and chairs. This allows our approach to pre-plan socially aware paths based on expected locations, with dynamic adjustments handled by complementary techniques. Despite this assumption, our work marks a significant step toward social navigation strategies for mobile manipulators, especially when transporting large objects that may cause discomfort.

        Our approach integrates all system components into the planning process using a risk-aware, sampling-based method, where the traditional risk cost is replaced by a social-related cost function. 
        The framework consists of three key steps: first, defining the social function; second, executing path planning with Risk-RRT*~\cite{Primatesta2020}, which integrates the social function and accounts for both the robot and the object being transported using a multi-point approach; and third, implementing path-following and control strategies.
        Results demonstrate that our method effectively generates paths that prioritize human comfort and safety without compromising operational efficiency. In this sense, our method outperforms current mobile-oriented social navigation methods.        

    The contributions of our work are summarized as follows:
    \begin{itemize}
        \item Introduction of multi-interest-point social cost function;
        \item Enhanced socially-aware whole-body motion planner;
        \item Complete framework for socially-acceptable object transportation in human-populated environments.        
    \end{itemize}

\section{Related Work}
    %\douglas[red]{Estou trabalhando nessa seção.}

    %The earliest works in social navigation can be traced back to mobile tour guide robots \cite{Mavrogiannis2023}, with the topic being widely studied in recent years \cite{Hirose2023, Silva2023, Dong2024}.    
    %Robotic social navigation is being widely studied in recent years \cite{Hirose2023, Silva2023, Dong2024}. However, although multiple approaches can be found for socially-aware mobile robots, there are far fewer developed specifically for mobile manipulators. This is reflected in the fact that the majority of commercial social robots do not have manipulators or only use the arm for gesture \cite{Mahdi2022}.

    Robotic social navigation continues to gain attention in recent years \cite{Hirose2023, Silva2023, Dong2024}. 
    %However, while several approaches have been developed for mobile robots, far fewer are tailored specifically for mobile manipulators . 
    Despite the growing deployment of mobile manipulators in human environments~\cite{Thakar2022}%~\cite{Garrell2017,Thakar2022}, 
    most socially-aware navigation research still focuses on the mobile platforms only or underutilizes the manipulator’s capabilities~\cite{Mahdi2022}.
    Social navigation in mobile robots has been explored using Reinforcement Learning~\cite{Xu2023, Narayanan2023}, control-based techniques~\cite{Poddar2023}, and RRT-based algorithms~\cite{Silva2023, Chi2018}. However, applying these methods to mobile manipulators presents challenges: control-based approaches require new architectures, and Reinforcement Learning must account for the manipulator’s influence, making them not directly transferable.
    
    RRT-based methods for mobile robots can be adapted by adding collisions with the manipulator, being well-suited for whole-body motion planning in high-dimensional spaces~\cite{Shao2021}.
    The work by Silva \etal~\cite{Silva2023} utilize RRT* and Informed RRT* algorithms in conjunction with a social heatmap to generate socially-aware navigation paths.
    Chi \etal~\cite{Chi2017} introduced a socially-aware modification of the Risk-RRT* algorithm, and later extended it to Risk-Informed-RRT*~\cite{Chi2018} for better performance.
    %Chi \etal~\cite{Chi2017} introduced the Risk-RRT algorithm, integrating a comfort and collision risk map into the RRT framework to facilitate socially-aware navigation. In subsequent years, they extended this work by developing Risk-RRT* and Risk-Informed-RRT*~\cite{Chi2018}, further enhancing the method's ability to account for both social comfort and safety during navigation.
    %
    While RRT-based methods are adaptable, they still only account for discomfort caused by the mobile base, neglecting the manipulator and any carried objects. This limitation, common to all current socially-aware planning methods, is addressed in our work.

    Regarding mobile manipulators, few works have appeared in the literature on social navigation.
    Sisbot \etal~\cite{Sisbot2007, Sisbot2007-H} introduced foundational approaches for socially-aware mobile manipulator navigation. In \cite{Sisbot2007}, they tackled a fetch-and-carry task using a mobile manipulator, considering factors such as safety, visibility, and arm comfort. However, their method treated navigation and manipulation as distinct tasks and focuses on interaction.
    %However, their method treated navigation and manipulation as distinct tasks: the robot first navigated to a desired location, and only after becoming stationary did the manipulator handle the object. 
    In a subsequent work \cite{Sisbot2007-H}, they extended the approach by introducing a human-aware motion planner for mobile manipulators, enabling the robot to pick up and manipulate objects in a socially acceptable manner. Nonetheless, the focus remained on navigation, with limited attention to manipulation or the object being transported.
    %However, even in this extended approach, the manipulator was largely stationary during operation.

    In our work, we extend a risk-based variant of Optimal-RRT (RRT*)~\cite{Primatesta2020, Chi2017}, mapping risk to social discomfort. Our proposed multi-interest point approach accounts for discomfort from the mobile base, manipulator, and transported object, enabling collision avoidance and full socially-aware navigation for a mobile manipulator. Discomfort is quantified using Hall’s proxemics concept~\cite{Hall1966}.

    To the best of our knowledge, no prior work simultaneously considers both the arm and mobile base in socially-aware navigation for mobile manipulators while allowing concurrent actuation. Moreover, our approach uniquely incorporates the discomfort caused by an object being transported, further enhancing its novelty.

\section{Problem Formulation}
%\section{Preliminaries} igual problem formulation?
 % World Modeling;
 % Static environment;
 % Robot and Object Kinematics

Let $\mathcal{E} \in \mathbb{R}^2$ be a known static cluttered and human-populated environment, represented by a map $\mathcal{M}$. The space occupied by obstacles is denoted as $\mathcal{M}_{obs}$, while the free space is defined as $\mathcal{M}_{free} = \mathcal{M} \backslash \mathcal{M}_{obs}$.
%\douglas{Adicionar a parte dos obstaculos/mapa.}

Consider a set of spatially distributed individuals denoted as $\peopleset = \{\individual{1}, \ldots, \individual{\numpersons}\}$.
An individual $\individual{i}$ is characterized by its configuration $\config{i} = (\pos_i, \theta_i)$, where $\pos_i = (x_i, y_i)$ denotes the position, and $\theta_i$ the orientation.

%The robot is a Mobile manipulator with a holonomic base platform

Let $\mathcal{R}$ be a holonomic mobile base represented by its position $\basepos{} = (x_\textrm{base}, y_\textrm{base})$, with kinematic model $\dot{\mathbf{p}}^\textrm{B} = u$. Without loss of generality, we assume the system includes an attached planar manipulator with two degrees of freedom.

The complete mobile manipulator configuration is:
\begin{equation}
    \mmconf{} = 
    \begin{bmatrix}
        x_\textrm{base} &
        y_\textrm{base} &
        \psi_1 &
        \psi_2        
    \end{bmatrix}^{\textrm{T}}
    ,
    \label{eq:mmconf}
    \nonumber
\end{equation}
where $\psi_1 , \psi_2 \in [0, 2\pi]$ are the manipulator joints angles.

%\douglas{Citar métrica ou custo que queremos otimizar. \caio{Seria exatamente o social cost não?}}

%The start and end configurations are $\mmconf{start}$ and $\mmconf{end}$ respectively, with $\mmconf{start},\mmconf{end} \in \mathcal{M}_{free}$ . The motion connecting the start and the end configuration is denoted as $\sigma(\mmconf{start},\mmconf{end})$. The optimal motion planner searches for an optimal set of configurations forming a path $\sigma^*(\mmconf{start},\mmconf{end})$ as a sequence of $\mmconf{}$ states from $\mmconf{start}$ to $\mmconf{end}$ which minimizes a given social cost function $\texttt{SocialCost}(\cdot)$. Hence, the optimal motion path is the solution of the following

Given $\mmconf{start}$ and $\mmconf{end}$, start and end configurations, with $\mmconf{start}, \mmconf{end} \in \mathcal{M}_{free}$. The path connecting the start and end configurations is represented by $\sigma(\mmconf{start}, \mmconf{end}) \in \mathcal{M}_{free}$. The optimal motion planner aims to determine an optimal sequence of configurations forming a path, $\sigma^*(\mmconf{start}, \mmconf{end})$, that minimizes a given cost function $\mathcal{F}(\cdot)$, where the cost is socially driven. Formally, the objective is to find the optimal solution to the following problem:
{
\begin{equation}
    \sigma^*(\mmconf{start},\mmconf{end}) = \underset{\sigma(\mmconf{start},\mmconf{end})}{\arg\min}  
    \mathcal{F}(\cdot) ~.
     \label{eq:minimization}
\end{equation}
}

%which minimizes {a given social cost function $\mathcal{S}(\cdot)$. Formally, we seek to find the optimal solution to the following problem:
%
% {\small
% \begin{equation}
%     \sigma^*(\mmconf{start},\mmconf{end}) = \underset{\sigma(\mmconf{start},\mmconf{end})}{\arg\min}  
%     \sum \mathcal{F}(\cdot) ~.
%      \label{eq:minimization}
% \end{equation}
% %
% }

% \douglas{Vamos deixar como $\mmconf{start}$ e $\mmconf{end}$, já que todos os demais nós estão por extenso. A gente dá um jeito depois de formatar a equação.\caio{feito}}

% The $\mmconf{end}$ and the social cost function \texttt{MotionSocialCost} used in this work will be defined in the following section.

% A high-level definition of the problem is:
% \begin{problem}[Social Object Transportation]
%     Given a known static cluttered and human-populated environment, a mobile manipulator needs to navigate and transport a non-negligible object in a safe, efficient, and socially acceptable manner.
% \end{problem}

\section{Methodology}

In this work, we propose a comprehensive framework for a mobile manipulator to navigate while transporting an object in a socially acceptable manner.

Our approach generates an integrated motion for the mobile platform, its attached manipulator, and the transported object, leveraging a specialized objective function that must be optimized by the planner. To achieve this, we extend a conventional risk-aware method~\cite{Primatesta2020} by substituting the traditional risk cost with a socially aware cost function, and considering multiple interest points across the robot’s base, manipulator, and object. This adaptation allows the planner to balance social acceptability, operational efficiency, and safety across all components of the mobile manipulator.

%The next sections outline our proposed methodology for socially-aware mobile manipulation. 
First, we introduce the social cost function, specifically designed for mobile manipulation tasks, which accounts for the full positioning of the entire system. Next, we introduce the Social Risk-RRT* motion planner, detailing its adaptation to incorporate social cost into the planning process. Lastly, we discuss the path-following strategy and the control structure implemented to execute the planned trajectories in simulation.

\subsection{Social cost function}

In order to achieve a socially-aware transportation, it is essential to quantitatively assess a person's level of comfort based on the motion occurring in their surroundings.
In this work, we build upon the concept of \emph{proxemics}, where we adopt the traditional Asymmetric Gaussian Function (AGF)~\cite{Kirby2010} as the personal space model. It will be employed to quantify a person’s comfort level based on the complete system's position (base, arm, object) relative to them.

Algorithm~\ref{alg:agf} can be used to determine the social cost of a specific coordinate $(x,y)$ on the plane relative to an individual $\individual{i}$, with higher values indicating increased discomfort. The closer the coordinate is to the person, the higher the associated social cost. Following the model proposed by Kirby~\cite{Kirby2010}, the AGF is centered at $\pos_i = (x_i, y_i)$, with orientation $\theta_i$, and specified variances $\sigma_h$, $\sigma_s$, and $\sigma_r$ along the person's direction, sides, and rear.

%\douglas{Se formos efetivamente mencionar as equações, melhor na forma de algoritmo.}

\begin{algorithm}[htpb]
    \caption{$\texttt{AGF}(x,y)$ \cite{Kirby2010}}
    \label{alg:agf}
    \begin{algorithmic}[1]
    \STATE $\alpha \gets \atantwo(y-y_i, x-x_i) - \theta_i + \pi/2$
    \STATE Normalize $\alpha$
    \STATE $\sigma \gets (\alpha \leq 0 ~?~ \sigma_r : \sigma_h)$
    \STATE $a \gets (\cos\theta_i)^2/(2\sigma^2) + (\sin\theta_i)^2/(2\sigma_s^2)$
    \STATE $b \gets \sin(2\theta_i)/(4\sigma^2) - \sin(2\theta_i)/(4\sigma_s^2)$
    \STATE $c \gets (\sin\theta_i)^2/(2\sigma^2) + (\cos\theta_i)^2/(2\sigma_s^2)$
    \RETURN {\small$ \exp(-(a(x-x_i)^2 + 2b(x-x_i)(y-y_i) + c(y-y_i)^2))$}
    \end{algorithmic}
\end{algorithm}

%Similar to other studies \cite{RiosMartinez2011Understanding,Vasquez2013Human, SilvaMacharet2019}, 
Similar to other studies \cite{Vasquez2013Human, SilvaMacharet2019}, which suggest that people are more protective of their frontal space, we model the AGF as elongated along the body orientation. 
%Specifically, we adjust the value of $\sigma_h$, defined by \cite{Kirby2010} as:
%
Specifically, we set the following parameters:
\begin{equation}
    \sigma_h = 2 \quad \textrm{and} \quad
    \sigma_s = \frac{2}{3} \sigma_h \quad \textrm{and} \quad
    \sigma_r = \frac{1}{2} \sigma_h ~.
    \nonumber
\end{equation}

%
%\begin{equation}
%    \sigma_h = \max(2v, 1/2) ~,
%    \nonumber
%\end{equation}
%
%to consider a constant linear velocity of $v = 1.0$m/s, even for static individuals. The other variances are given by:
%
%\begin{equation}
%    \sigma_s = \frac{2}{3} \sigma_h \quad \textrm{and} %\quad
%    \sigma_r = \frac{1}{2} \sigma_h ~.
%    \nonumber
%\end{equation}

Additionally, we apply a threshold of $\tau = 0.2$ to evaluate the AGF limited to the proxemic zone known as \emph{public space}~\cite{Patompak2016}, which extends up to approximately $3.60$m in a straight line from the person's front at its farthest point. If the AGF result is less than or equal to 0.2, it is considered 0, indicating no discomfort.

While the use of social functions has been successfully applied to guide mobile robots in socially acceptable navigation \cite{Chi2017,Silva2023,Patompak2016}, additional considerations are required for mobile manipulators. Traditional methods typically focus on the robot’s mobile base as the key reference point. However, when applied directly to mobile manipulators, these methods may prevent the base from entering socially restricted areas, yet fail to account for the manipulator arm or the carried object, which can still intrude into personal space.

In this context, we propose a multi-interest-point approach to quantify the social acceptability of the robot based on its complete current configuration.
Additionally, it allows for assigning different weights to various points, giving priority to those that are most critical. For instance, if the transported object is hazardous, higher weights can be assigned to the object's key points, emphasizing the importance of keeping it farther from people. %In such cases, a robot's base being closer to a person is more socially acceptable than having the dangerous object near them.

To implement this approach, key points on the robot must be defined to calculate the social cost, with these points being selected based on the robot’s geometry. Intuitively, corners and tips are suitable choices, as they represent critical areas. For instance, on a planar arm with revolute joints, the tip of a link can serve as a key social point -- if the tip is far from a person, it is likely that the entire arm remains distant as well. 
Moreover, corners and tips often pose higher risks due to their sharper features, making them important for social cost evaluation.  
The same principle applies to the object being carried, where extremities are logical points for defining social importance. These points are manually selected, future works could explore automatic selection methods.

\begin{figure}[htbp]
    \centering
    \begin{subfigure}{0.40\linewidth}
        \centering
        \includegraphics[width=\linewidth, trim={0.75cm 0cm 1.25cm 0.1cm}]{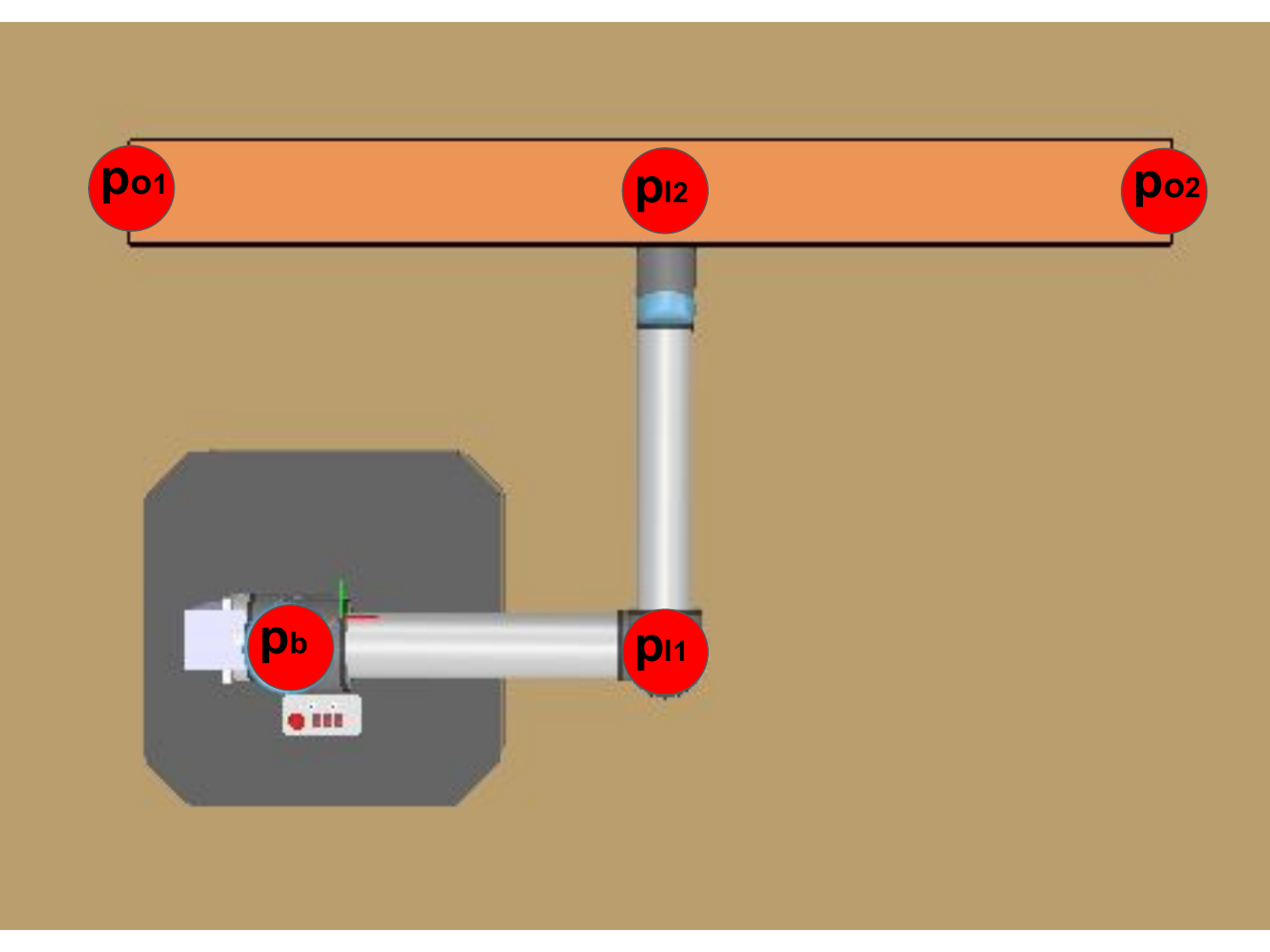}
        \caption{Bar-shaped object.}
        \label{fig:o1-points}
    \end{subfigure}
    \hspace{2mm}
    %\hfill
    \begin{subfigure}{0.40\linewidth}
        \centering
        \includegraphics[width=\linewidth, trim={0.75cm 0cm 1.25cm 0.1cm}]{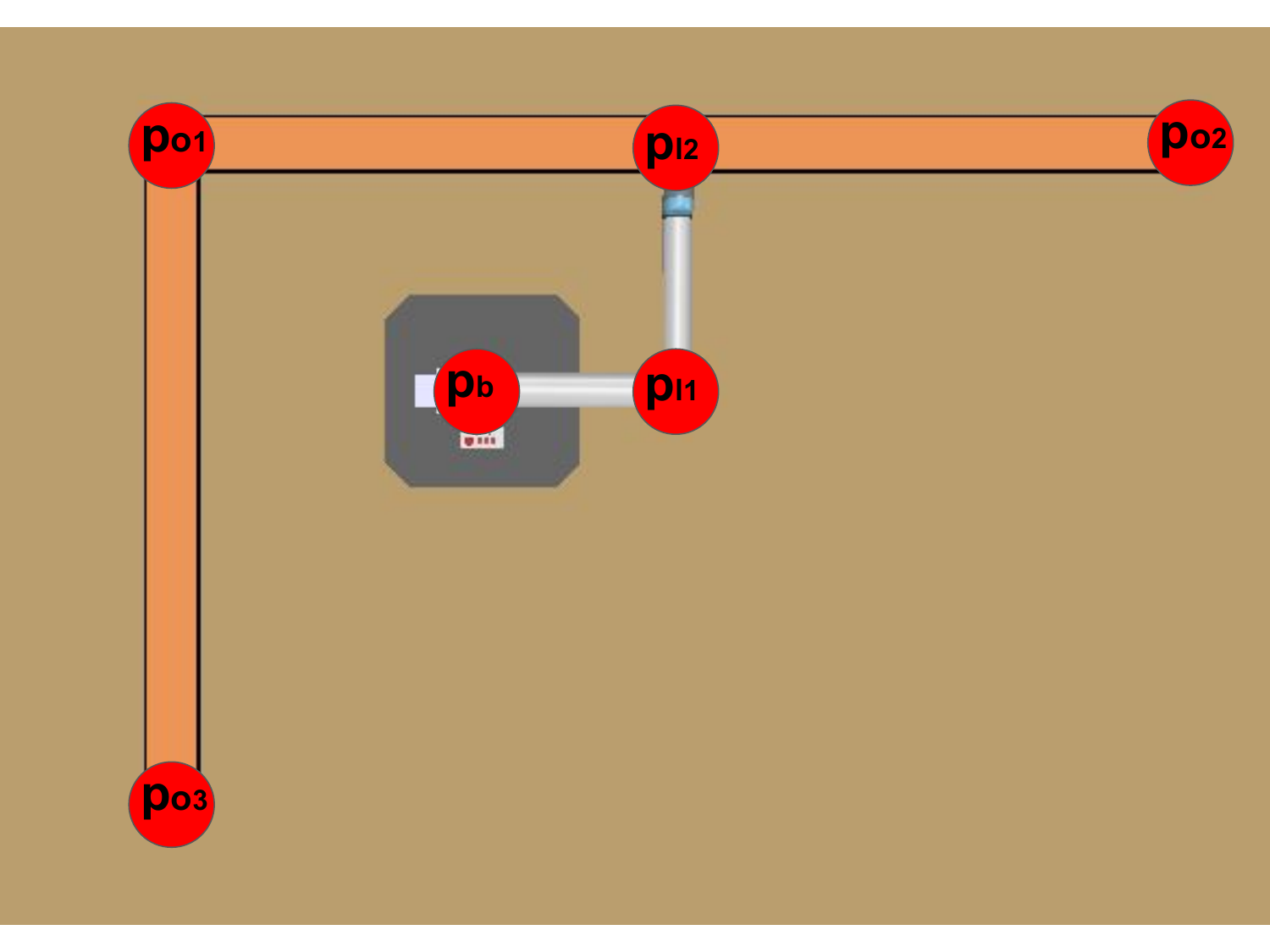}
        \caption{L-shaped object.}
        \label{fig:o2-points}
    \end{subfigure}
    \caption{%The points $p_b$, $p_{l1}$, $p_{l2}$, $p_{o1}$, $p_{o2}$, and $p_{o3}$ (only present in (b)) correspond to the base, tip of the first link, tip of the second link, first tip of the object, second tip of the object, and the last tip of the L-shaped object, respectively.
    The points $p_b, p_{l1}, p_{l2}, p_{o1}, p_{o2}$, and $p_{o3}$ (only in (b)) represent the base, link tips, and object tips, including the final tip of the L-shaped object.
    }
    \label{fig:interest-points}
\end{figure}

In this work we consider a 2-link planar mobile manipulator with revolute joints carrying an object; examples of defined points for different objects are shown in Figure~\ref{fig:interest-points}.

%The social cost $S_r$ of a robot relative to $k$ people can be calculated as the sum of the weighted social costs for all defined $n$ robot points relative to each person. Specifically,For each person $j$ the social cost $S_j$ is computed as the weighted sum of the social costs at $n$ robot points, where $w_i$ represents the weight of each point and $F(x_i,y_i)$ is the cost social function at point $(x_i,y_i)$ presented in Equation \ref{eq:individual-social-func}. Therefore, the overall social cost $S_r$ is given by:

Given $n$ selected points, the total social cost of a mobile manipulator at configuration $\mmconf{}$ relative to all people $h$ is calculated as a weighted sum of the social costs of each individual point/person, \ie:
\begin{equation}
    %\texttt{SocialCost}(\cdot) = 
    \mathcal{S}(\cdot) =
    \sum_{i=1}^{\numpersons} \sum_{j=1}^{n} w_j \texttt{AGF}_i(x_j, y_j),
     \label{eq:social-func-weight}
\end{equation}
where $w_j$ is the point's weight, and \texttt{AGF} is the social cost related to an individual's personal space, defined by Alg.~\ref{alg:agf}. %\rev{ As explained, the weights need to be tuned based on the task, object, and manipulator, tuning these parameters may take some time to achieve the desired results.}

\subsection{Motion planning}
%\rmv{Sampling-based methods are often chosen for motion planning in complex scenarios, particularly for problems with a high number of degrees of freedom, such as mobile manipulation. Unlike traditional methods, which can become computationally infeasible with increasing complexity, sampling-based techniques efficiently explore the space by randomly sampling and connecting feasible paths.}

In this work, we adapt the Risk-based Optimal Rapidly-exploring Random Tree (Risk-RRT*) by \cite{Primatesta2020} method, incorporating a social cost function as defined by Equation~\ref{eq:social-func-weight}, instead of the traditional risk function.

The pseudocode for our Social Risk-RRT* is reported on Algorithm~\ref{alg:RiskRRT}, where the algorithm mainly differs from the original one in Line 14, where the social costs are introduced. 
%The rewiring proccess is reported on Algorithm \ref{alg:rewire}, differing from RRT* to consider the social cost instead of euclidean distance.
%
%The Social Risk-RRT* proposed in this paper, as all RRT-based algorithms, constructs a Tree $T$ formed by Nodes $p$, each node has one parent $p.parent$. This is also commonly described and used as a graph.

\algsetup{indent=1em}
\begin{algorithm}[htpb]
    \caption{Social Risk-RRT*} \label{alg:RiskRRT}  
    
    \begin{algorithmic}[1]
    % \KwIn{\textit{Map}, $p_{goal}$, $p_{start}$}   
    % \KwOut{Nodes}
    \STATE $\mathcal{T} \gets \{\mmconf{start}\}$
    
    \FOR {$k = 1$ \TO $K$}    
        \STATE $\mmconf{rand} \gets \texttt{SampleRandomState}()$
        \STATE $\mmconf{nearest} \gets \texttt{NearestNode}(\mmconf{rand}, \mathcal{T})$
        
        \IF{$||\mmconf{nearest} - \mmconf{rand}|| > \delta$}
            \STATE {$\mmconf{new} \gets \texttt{Steer}(\mmconf{rand},\mmconf{nearest},\delta)$}
        \ELSE
            \STATE {$\mmconf{new} \gets \mmconf{rand}$}
        \ENDIF

        \IF{\texttt{isValid$\left(\mmconf{new}\right)$}}   
            \STATE $c_{min} \gets \infty$
            %\STATE $SC \gets \texttt{SocialCost}(p_{new})$ 
            \STATE  $\mathcal{Q}_{near} \gets \texttt{AllNear}(\mathcal{T}, \mmconf{new}, r_{near})$
            
            \FOR{ \textbf{each} $\mmconf{near} \in \mathcal{Q}_{near}$}
                \STATE $\Phi \gets \texttt{Msc}(\mmconf{near},\mmconf{new}, \mathcal{M})$
                
                \IF {$\mathcal{F}(\mmconf{near})$$ + \Phi < c_{min}$}
                    \IF{\texttt{CollisionFree($\mmconf{near}, \mmconf{new}, \mathcal{M}$)}}
                            \STATE $\mmconf{new}$.parent $\gets \mmconf{near}$
                            \STATE $c_{min}$ $\gets\mathcal{F}
                            (\mmconf{near})$ $+ \Phi$
                            
                            %\STATE $d_{min} \gets p_{near}.cost + \Phi$
                            %\STATE $p_{new}.parent \gets p_{near}$
                            %\STATE $p_{new}.cost \gets d_{min}$
                    \ENDIF 
                \ENDIF 
            \ENDFOR
       
            \IF{$c_{min} \neq \infty$}
                \STATE $\mathcal{T} \gets \mathcal{T} \cup \mmconf{new}$
                \STATE $\mathcal{T} \gets \texttt{Rewire}(\mathcal{T}, \mmconf{new}, \mathcal{Q}_{near},\mathcal{M})$
            \ENDIF 

        \ENDIF
        
    \ENDFOR

    \RETURN \texttt{GetPath}($\mathcal{T}$, $\basepos{end}$)
    
    \end{algorithmic}   
\end{algorithm}

%The algorithm begins by inserting in an empty tree $\mathcal{T}$ the first node $\mmconf{start}$, representing the robot's start configuration. It then iteratively runs for $K$ iterations. In each iteration, it samples node $p_{rand}$ uniformly randomly from the C-space and finds the nearest node $p_{nearest}$ in $T$ based on the norm.
%
%Given a maximum step size/planning range $\delta$, the algorithm checks if $p_{nearest}$ is within $\delta$ of $p_{rand}$. If is, the new node $p_{new}$ is set to $p_{rand}$, otherwise, a new node $p_{new}$ is created from $p_{nearest}$ in the direction of $p_{rand}$, at a distance of $\delta$, using the \texttt{Steer()} function as detailed in Equation \ref{eq:steer}. This initial part is equal to classical RRT* algorithms.

The algorithm initializes by inserting the start configuration, $\mmconf{start}$, as the first node into an empty tree $\mathcal{T}$. It then proceeds iteratively for $K$ iterations. In each iteration, a random node $\mmconf{rand}$ is uniformly sampled from the configuration space (C-space), and the nearest node $\mmconf{nearest}$ in the tree is determined based on the chosen distance metric. In this work, distances are computed as the norm of the difference between two configurations.
Given a planning range $\delta$, the algorithm first checks whether $\mmconf{nearest}$ lies within a distance $\delta$ from $\mmconf{rand}$. If this condition is met, the new node $\mmconf{new}$ is set to $\mmconf{rand}$; otherwise, a new node is created by moving from $\mmconf{nearest}$ toward $\mmconf{rand}$, at a distance of $\delta$, using the following function:
\begin{equation}
    \texttt{Steer}\left(\mmconf{1},\mmconf{2},\delta\right) = \mmconf{1} + \left( \frac{\mmconf{2} - \mmconf{1}}{||\mmconf{1}, \mmconf{2}||} \right)\delta ~.
    \label{eq:steer}
    \nonumber
\end{equation}

At Line 10, the algorithm checks whether $\mmconf{new}$ is valid, ensuring it does not collide with any obstacles or people. In Line 12, all nodes within a distance $r_{near}$ of $\mmconf{new}$ are put into set $\mathcal{Q}_{near}$. These nearby nodes are then used to determine the most suitable parent for $\mmconf{new}$.

%At line 11, the algorithm verifies if $p_{new}$ is valid, i.e. if is not colliding with any obstacles or people. In line 12, all nodes with norm less than $r_{near}$ from $p_{new}$ are stored into the set $P_{near}$. These nodes are used to determine the parent for $p_{new}$.

%In lines 14 to 19, the parent node for $p_{new}$ is selected by evaluating each node in $P_{near}$. The selection is based on the minimum motion social cost and if the motion between those two nodes is collision-free. The \texttt{MotionSocialCost()} function calculates the total social cost of moving from a potential parent node to $p_{new}$ (line 15). This involves linearly interpolating between the configurations of $p_{near}$ and $p_{new}$, and computing $S_r$ for each configuration, adding the results to obtain the motion social cost. 
%
%To ensure the path from $p_{nearest}$ to $p_{new}$ is feasible, the algorithm checks for collisions using the \texttt{CollisionFree()} function (line 16), the function linearly interpolate configurations between $p_{near}$ and $p_{new}$, verifying that none of these configurations collide with obstacles or people. 

In Lines 13 to 21, the parent node for $\mmconf{new}$ is chosen by evaluating each node in $\mathcal{Q}_{near}$. The selection criterion minimizes the cost from the start node to $\mmconf{new}$, using:
\begin{equation}
   \mathcal{F}(\mmconf{a}) = 
   \mathcal{F}(\mmconf{ap}) + 
   \texttt{Msc}(\mmconf{ap},\mmconf{a})  ~,
\end{equation}
%
%\rev{where $\mmconf{ap}$ is the parent of $\mmconf{a}$ and $\texttt{Msc}(\cdot)$ is the motion social cost. The motion social cost computes the total social cost of moving from one configuration to another. In Line 14, it computes the total social cost of moving from a potential parent node to $\mmconf{new}$. Similar to Primatesta \etal~\cite{Primatesta2020}, this is done by integrating the social cost over the distance moved $s$:}
where $\mmconf{ap}$ is the parent of $\mmconf{a}$ and $\texttt{Msc}(\cdot)$ computes the motion social cost, which quantifies the total social cost of moving between configurations. In Line 14 it is calculated the social cost of moving from a potential parent node to $\mmconf{new}$ by integrating the cost over the moved distance $s$, as done by Primatesta \etal~\cite{Primatesta2020}, \ie:

%
%The \texttt{Msc}$(\cdot)$ function computes the total social cost of moving from a potential parent node to $\mmconf{new}$ (Line 14).
%This calculation involves linearly interpolating between the configurations of $\mmconf{near}$ and $\mmconf{new}$, computing the social cost as in Eq.~\eqref{eq:social-func-weight} at each interpolated configuration, and integrating these values to determine the overall motion social cost.
%We show an illustration of the linear interpolation in Figure~\ref{fig:interpolation-example}, with their calculated values in Table~\ref{tab:table-interpolation}.
%
\begin{equation}
   \texttt{Msc}(\mmconf{a},\mmconf{b}) = \int_{\mmconf{a}}^{\mmconf{b}}{\mathcal{S}(\mmconf{})\text{d}s} ~.
\end{equation}

\begin{figure}[htpb]
    \centering
        \includegraphics[width=0.5\linewidth, trim={0.25cm 1.5cm .40cm 1cm}]{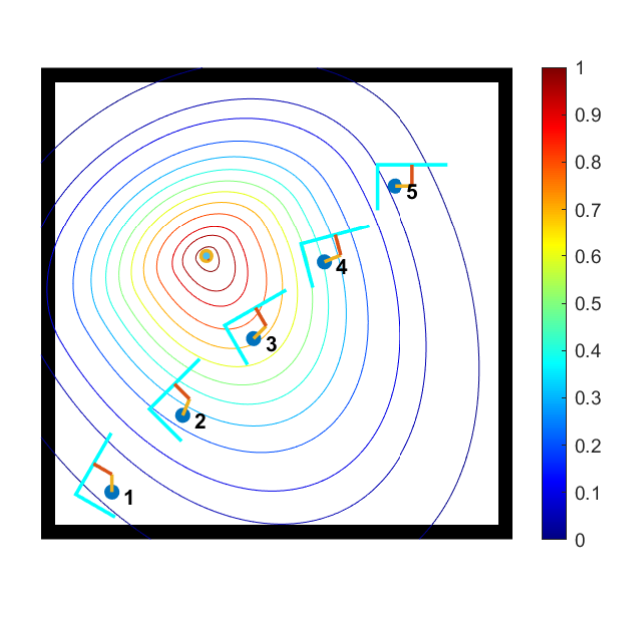}
        %\vspace{-7mm}    
        \caption{%\rmv{Linear interpolation example with one person but no social navigation for illustration purposes. The dark blue circle represents the robot's base, the yellow rod is the first link, and the orange rod is the second link. The light-blue L-structure is the object being carried. The robot begins at 1 and aims to reach the configuration 5. With a linear interpolation of three steps, intermediate configurations at 2, 3, and 4 are generated. The colormap represents the social cost.}
        Linear interpolation without social awareness: The dark blue circle represents the robot's base, the yellow rod is the first link, and the orange rod is the second link. The light-blue L-structure is the object. Starting from configuration 1, the robot reaches configuration 5 through intermediate steps 2, 3, and 4. The colormap represents the social cost.
        }
        \label{fig:interpolation-example}
\end{figure}

In practice, the integral is computed numerically by linearly interpolating between the start and end configurations, then applying the trapezoidal method\footnote{\url{https://www.mathworks.com/help/matlab/ref/trapz.html}}, as:
\begin{equation}
    \hspace{-2mm}
    \resizebox{.92\linewidth}{!}{
    $
    \begin{aligned}  
        \int_{\mmconf{a}}^{\mmconf{b}}{\mathcal{S}(\mmconf{})\text{d}s}         \approx
        \frac{1}{2} 
        \sum_{j=2}^{N} 
        ||\mmconf{j} -\mmconf{j-1}|| 
        \left[ \mathcal{S}(\mmconf{j}) + \mathcal{S}(\mmconf{j-1}) \right] ~,
    \end{aligned}
    $
    }
\end{equation}
where $\mmconf{a} = \mmconf{1}$, $\mmconf{b} = \mmconf{N}$ and intermediate configurations are the linear interpolation of $N$ steps between \mmconf{a} and \mmconf{b}.
%
%\rmv{We show an illustration of the linear interpolation in Figure~\ref{fig:interpolation-example}, with their calculated values in Table~\ref{tab:table-interpolation}.
%}
%
Figure~\ref{fig:interpolation-example} illustrates the linear interpolation step.

\vspace{2mm}

To ensure the feasibility of the path from $\mmconf{near}$ to $\mmconf{new}$, the algorithm utilizes the \texttt{CollisionFree}$(\cdot)$ function (Line 16). This function once again utilizes the linear interpolation between $\mmconf{near}$ and $\mmconf{new}$, checking each interpolated configuration to ensure it does not collide with any obstacles or people. %Only if all intermediate configurations are collision-free is the path deemed valid.
In Lines 17 and 18, the best parent node and its corresponding cost are recorded.

In Lines 22 to 25, if a valid $\mmconf{new}$ with a parent is found, the tree is \emph{rewired}, and $\mmconf{new}$ is added to the tree. During the rewiring process, each node in $\mathcal{Q}_{near}$ is evaluated to see if $\mmconf{new}$ can serve as a better parent. If using $\mmconf{new}$ as a parent results in a collision-free motion and reduces the current $\mathcal{F}(\mmconf{near})$, the parent of $\mmconf{near}$ is updated to $\mmconf{new}$. This step ensures the path is continuously optimized by incorporating lower-cost connections where possible.

%After all $m$ iterations, the algorithm retrieves the path using the \texttt{GetPath()} function (line 23). The function needs to identify the node closest to the goal configuration $p_{goal}$, then reconstructs the path by recursively tracing back through the parent nodes, incrementally constructing the path from the goal to the start configuration.

Finally, after completing all $K$ iterations, the algorithm retrieves the path using the \texttt{GetPath}$(\cdot)$ function (Line 28). This function first identifies the end node $\mmconf{end}$, which is the node within $r_{near}$ distance of the goal with the lowest total cost $\mathcal{F}(\cdot)$, and then reconstructs the path by recursively tracing back through the parent nodes. 
%\rmv{This process incrementally builds the path in reverse, from the goal to the start configuration, ensuring that the entire path is feasible and optimized.}

As we work in static environments, the algorithm runs offline to generate a path. While similar methods have been used online for mobile robots~\cite{Chi2017,Silva2023}, they are not fast enough for mobile manipulators due to increased configuration complexity and collision checks. Hence, further adaptation is needed to enable real-time use.

\subsection{Path following and control}

The robot follows the path generated by the planner using a straightforward waypoint-following strategy. Each waypoint specifies a configuration with the variables ${x_\textrm{base}, y_\textrm{base}, \psi_1, \psi_2}$. These values represent the desired state of the robot at each waypoint, defining both the position of the mobile base and the joint orientations of the manipulator.

We control the mobile base and manipulator independently but simultaneously. The mobile base, equipped with omnidirectional four-wheel drive, is managed using a proportional controller with a gain of $K_p = 1.5$:
\begin{equation}
\begin{bmatrix}
v_x \\
v_y
\end{bmatrix}
= K_p 
\begin{bmatrix}
x_{\text{current}} - x_{\text{next}} \\
y_{\text{current}} - y_{\text{next}}
\end{bmatrix} ~, 
\end{equation}
and this controller regulates the velocities of the four wheels: 
\begin{equation}
    \begin{bmatrix}
        v_1 & v_2 & v_3 & v_4
    \end{bmatrix}^\mathrm{T}
    = \frac{1}{r} 
    \begin{bmatrix}
        1 & 1 & -1 & -1 \\
        -1 & 1 & 1 & -1
    \end{bmatrix}^\mathrm{T}
    \begin{bmatrix}
        v_x \\
        v_y
    \end{bmatrix} ~,
\end{equation}
%
%
% %
% \begin{equation}
% \begin{bmatrix}
% v_1 \\
% v_2 \\
% v_3 \\
% v_4
% \end{bmatrix}
% = \frac{1}{r} 
% \begin{bmatrix}
% 1 & -1 \\
% 1 & 1 \\
% -1 & 1 \\
% -1 & -1
% \end{bmatrix}
% \begin{bmatrix}
% v_x \\
% v_y
% \end{bmatrix} ~,
% \end{equation}
%
%where $r$ is the radius of the wheels, $v_x$ and $v_y$ are the linear velocities in the $x$ and $y$ directions, $x_{current}$ and $y_{current}$ are the current base coordinates $x_{base}$, $y_{base}$, and $x_{next}$, $y_{next}$ are the coordinates of the next waypoint to be reached by the base.
where $r$ is the radius of the wheels, $v_x$ and $v_y$ represent the linear velocities in the $x$ and $y$ directions, $x_{\text{current}}$ and $y_{\text{current}}$ are the current coordinates of the mobile base, while $x_{\text{next}}$ and $y_{\text{next}}$ are the coordinates of the next waypoint.
 
%\textcolor{red}{The manipulator, on the other hand, relies on pre-existing low-level control in the simulator, where we directly input the desired joint angles.}
%
The manipulator, on the other hand, relies on an implementation of the Ruckig online trajectory generator~\cite{berscheid2021jerk}\footnote{\url{https://github.com/pantor/ruckig}}, where we directly input the desired joint angles.

The robot only proceeds to the next waypoint once both the base and manipulator have reached the current waypoint configuration within a defined acceptance distance. This ensures precise alignment and coordinated movement before transitioning to the next segment of the path.

%\douglas{A metodologia não deve depender da seção de experimentos. Será que a gente consegue descobrir o controlador ou apenas citar um possível? \caio{De acordo com o coppelia usa: Ruckig online trajectory generator}}

% \douglas{Talvez adicionar alguma equação/algoritmo simples só para dar uma cara melhor na seção? \caio{adicionei da base}}

\section{Experiments}
%\rev{To evaluate our method, we first compare social algorithms to a non-social baseline (RRT*) in two environments (Sections A and B). Section C presents an in-depth comparison between our approach and other social navigation algorithms. Finally, Section D analyzes the impact of varying social cost weights. Each algorithm for each environment was executed 10 times with $K = 2000$. Except when specified, all multi-point approaches uses uniform weights assigned to all points ($\{w\}_{j=1}^5 = 1$).}

To evaluate our method, we first compare social navigation algorithms with a non-social baseline (RRT*) in two distinct environments (Sections \ref{sec:a} and \ref{sec:b}). Section \ref{sec:c} provides a detailed comparison between our approach and other social navigation methods, while Section \ref{sec:d} examines the effect of varying social cost weights on the resulting trajectories. Each algorithm was executed 10 times per environment with $K = 2000$. Unless stated otherwise, all multi-point approaches use uniform weights ($\{w\}_{j=1}^5 = 1$).

% \rev{
% For these experiments, we consider four social algorithms. Our main multi-point social Risk-RRT* approach, accounting for the full mobile manipulator. Then, we combine this version with the Informed-RRT*~\cite{Gammell2014}, inspired by its recent use in social navigation~\cite{Silva2023,Chi2018}, where sampling is restricted to an ellipsoid in the $xy$-plane after the first solution is found. And lastly, since existing approaches focus solely on the base, we compare both algorithms with their correspondent base-only variant, implemented by assigning zero weight to the manipulator and object in the social cost function and one to the base. %In summary, we evaluate: traditional RRT*, full and base-only Risk-RRT*, and full and base-only Risk-Informed-RRT*.
% }
%To evaluate our method, we performed experiments across various environments, object types, and weights. For each setting, we executed the method 10 times with $K = 2000$ and selected the best result, which is defined as the one with the lowest cumulative social cost for the given conditions.

For these experiments, we evaluate four social navigation algorithms. First, our multi-point social Risk-RRT* approach, which considers the full mobile manipulator, including the base, arm, and carried object. Next, we extend this approach with the Informed-RRT*~\cite{Gammell2014}, inspired by its recent applications in social navigation~\cite{Silva2023,Chi2018}, where sampling is restricted to an ellipsoid in the XY-plane after the first solution is found. Finally, since most existing methods focus solely on the base, we compare both algorithms with their corresponding base-only variants, implemented by assigning zero weight to the manipulator and object while keeping a weight of one for the base in the social cost function.

\subsection{Complex environment}
\label{sec:a}

\begin{figure}[htbp]
    \centering
    \begin{subfigure}{0.47\linewidth} % Adjust the width to fit in a single column
        \centering
        \includegraphics[width=\linewidth, trim={0.75cm 0cm 1.25cm 0.1cm}]{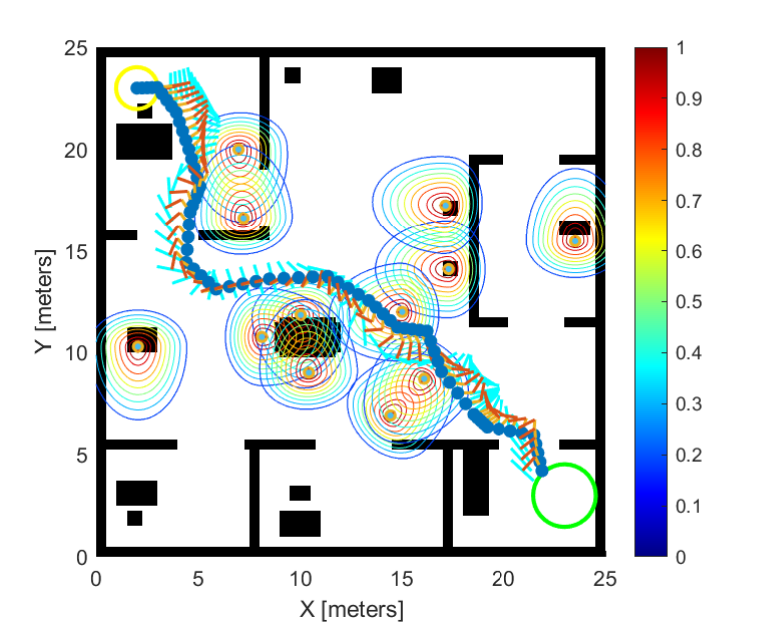} % Replace with your figure file
        \caption{Traditional RRT*.}
        \label{fig:complex-ns}
    \end{subfigure}
    \hspace{1mm}
    %\hfill
    \begin{subfigure}{0.47\linewidth} % Adjust the width to fit in a single column
        \centering
        \includegraphics[width=\linewidth, trim={0.75cm 0cm 1.25cm 0.1cm}]{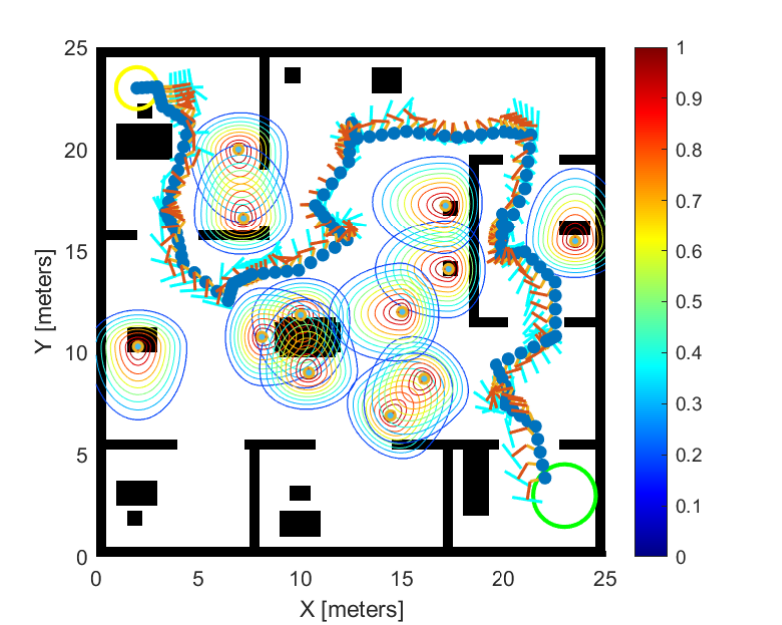} % Replace with your figure file
        \caption{Social Risk-RRT* (ours).}
        \label{fig:complex-s}
    \end{subfigure}
    % \begin{subfigure}{0.47\linewidth} % Adjust the width to fit in a single column
    %     \centering
    %     \includegraphics[width=\linewidth, trim={0.75cm 0cm 1.25cm 0.1cm}]{img/OfficePartial.pdf} % Replace with your figure file
    %     \caption{}
    %     \label{fig:complex-s}
    % \end{subfigure}
    \caption{Comparison of navigation results in an office-like environment, robot transporting a bar object, and all weights uniformly set to 1. Video: \href{https://youtu.be/lLFQ12lf\_go}{https://youtu.be/lLFQ12lf\_go}.}
    \label{fig:complex-comparison}
\end{figure}

\begin{figure}[htpb]
    \centering
    \includegraphics[width=0.6\linewidth, trim= 0 0 0 0, clip]{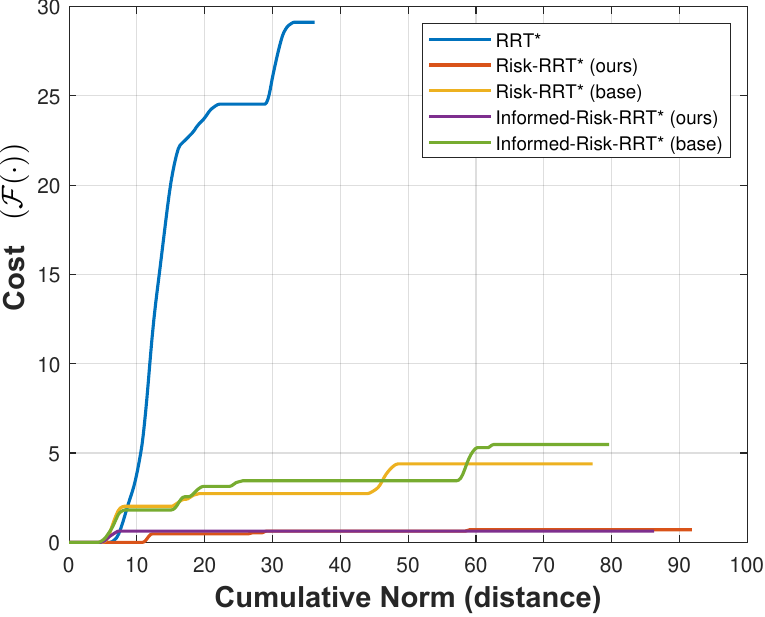}
    \caption{Graph comparing the results of social methods with RRT* in an office-like environment with a bar-shaped object. The Y-axis represents the cost, while the X-axis shows the cumulative norm.}
    \label{fig:graph-complex}
\end{figure}

The first environment is a complex office-like setting, where the robot transports a 1.5-meter-long bar (Figure~\ref{fig:o1-points}). Due to space constraints, we visualize only the best results from traditional RRT* and our Social Risk-RRT* in Figure~\ref{fig:complex-comparison}, as they provide a meaningful comparison of social versus non-social planning. A detailed analysis of social cost versus distance for all algorithms is presented in Figure~\ref{fig:graph-complex}.

%The first environment is a complex office-like setting, where the object to be transported is a 1.5-meter-long bar (Fig.~\ref{fig:o1-points}) with uniform weights assigned to all points ($\{w\}_{j=1}^5 = 1$). We plot the best results obtained using the traditional RRT* and our proposed method, shown in Fig.~\ref{fig:complex-comparison}, only those two are plotted due to space constraints. The cumulative social cost for all algorithms is detailed in Fig.~\ref{fig:graph-complex}.

Figure~\ref{fig:complex-comparison} shows that our approach effectively directs both the robot’s base and manipulator away from high-discomfort regions, reducing social disturbances. 
This trend is further supported by Figure~\ref{fig:graph-complex}, which shows that our method, along with other social approaches, consistently achieves lower social costs compared to RRT*.
However, as expected, our paths tend to be longer, since the RRT* minimize distance, while our approach optimizes for social acceptability.

%The results demonstrate that our method effectively guides the robot (base and manipulator) to avoid social regions, minimizing disturbances, in contrast to the RRT*. The graph in Figure~\ref{fig:graph-complex} supports this observation, showing that our method consistently maintains a significantly lower cumulative social cost throughout the path compared to the RRT*. However, it is worth noting that our path tends to be longer.

\subsection{Complex object}
\label{sec:b}

%\rev{In this section, we evaluated our method in a generic map and a L-shaped object, consisting of two bars measuring 3.0 meters and 2.0 meters in length (Fig.~\ref{fig:o2-points}). The environment is tighter and the object complex, increasing the difficulty of finding an optimal social path. The best results for both RRT* and our Risk-RRT* are shown in Fig.~\ref{fig:simple-comparison}, with the cost \textit{vs} distance graph for all algorithms displayed in Fig.~\ref{fig:graph-simple}.
%}
%In this section, we evaluated our method with the mobile manipulator carrying an L-shaped object, consisting of two bars measuring 3.0 meters and 2.0 meters in length (Fig.~\ref{fig:o2-points}). The environment is simpler but tighter, increasing the difficulty of finding an optimal social path, particularly given the complexity of the object. The best results for both RRT* and our method are shown in Fig.~\ref{fig:simple-comparison}, with the corresponding cumulative social cost displayed in Fig.~\ref{fig:graph-simple}.

\begin{figure}[htbp]
    \centering
    \begin{subfigure}{0.47\linewidth} % Adjust the width to fit in a single column
        \centering
        \includegraphics[width=\linewidth, trim={0.75cm 0cm 1.25cm 0.1cm}]{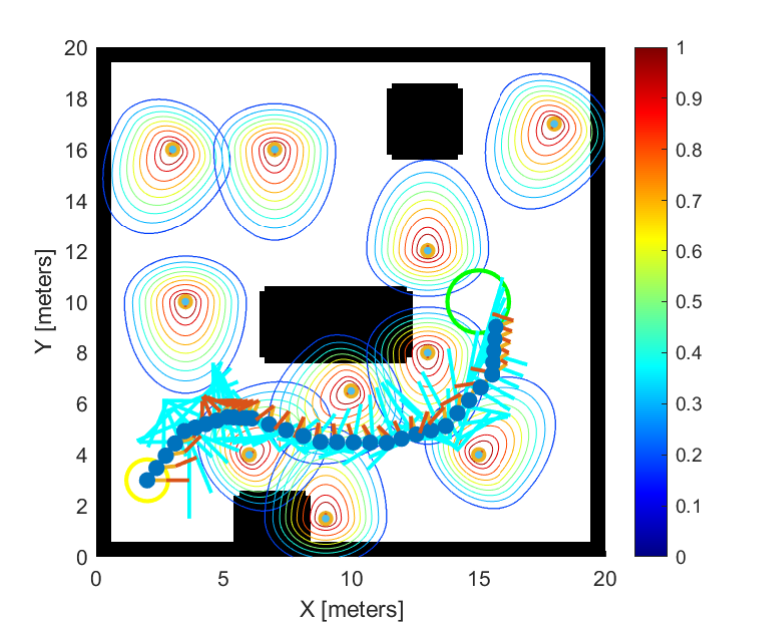} % Replace with your figure file
        \caption{Traditional RRT*.}
        \label{fig:simple-ns}
    \end{subfigure}
    \hspace{1mm}
    %\hfill
    \begin{subfigure}{0.47\linewidth} % Adjust the width to fit in a single column
        \centering
        \includegraphics[width=\linewidth, trim={0.75cm 0cm 1.25cm 0.1cm}]{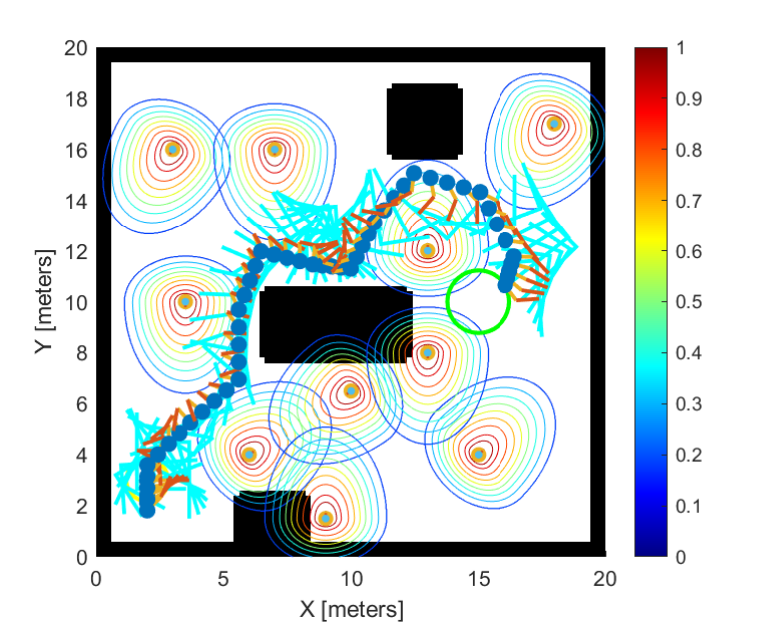} % Replace with your figure file
        \caption{Social Risk-RRT* (ours).}
        \label{fig:simple-s}
    \end{subfigure}
    % \begin{subfigure}{0.47\linewidth} % Adjust the width to fit in a single column
    %     \centering
    %     \includegraphics[width=\linewidth, trim={0.75cm 0cm 1.25cm 0.1cm}]{img/M1Partial.pdf} % Replace with your figure file
    %     \caption{}
    %     \label{fig:simple-s}
    % \end{subfigure}
    \caption{Comparison of navigation results in tighter environment, robot carrying a L-shaped object, and all weights uniformly set to 1. Video: \href{https://youtu.be/c8PtqI2iLMM}{https://youtu.be/c8PtqI2iLMM}.}
    \label{fig:simple-comparison}
\end{figure}

\begin{figure}[htpb]
    \centering
    \includegraphics[width=0.6\linewidth]{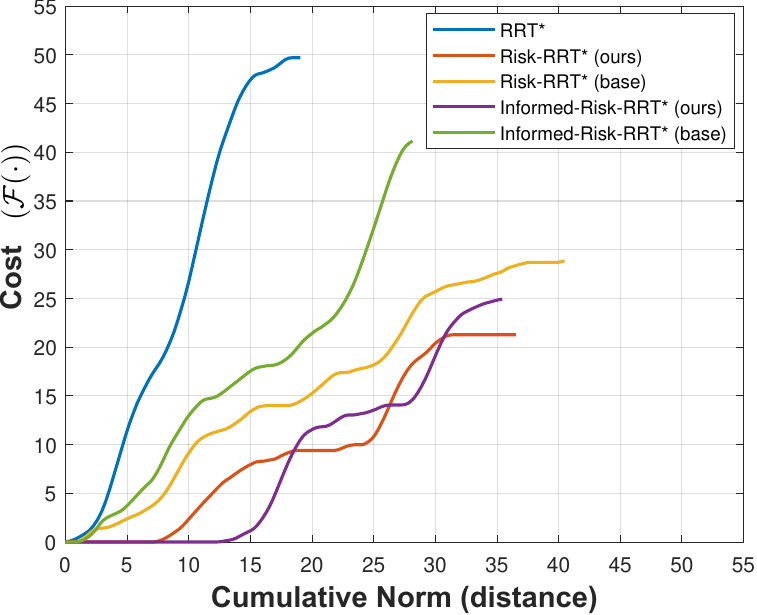}
    \caption{Graph comparing the results of social methods with RRT* in a tighter generic environment with a L-shaped object. The Y-axis represents social cost, while the X-axis shows the cumulative norm.}
    \label{fig:graph-simple}
\end{figure}

In this section, we evaluate our method in a generic map with a complex L-shaped object composed of two bars measuring 3.0 and 2.0 meters in length (Fig.~\ref{fig:o2-points}). The tighter environment and intricate object geometry make finding an optimal social path more challenging. Figure~\ref{fig:simple-comparison} presents the best results for both RRT* and our Social Risk-RRT*, while Fig.~\ref{fig:graph-simple} displays the cost \textit{vs.} distance graph for all algorithms.

% \rev{The results show that our method effectively navigates the robot while minimizing proximity to people when feasible. However, in narrow corridors, the robot is compelled to move closer to people due to restricted path choices. When comparing RRT* with our approach, RRT* tends to pass through the center of areas densely populated with people, while our method avoids these regions, producing a more socially-aware path. This is supported by the graph in Fig.~\ref{fig:graph-simple}, which highlights that, despite a longer path, the cost is consistently lower for all social methods when compared to the RRT*.}
%Despite the described challenges, our approach proved successful. The results show that our method effectively navigates the robot through tight spaces while minimizing proximity to people when feasible. However, in narrow corridors, the robot is compelled to move closer to people due to restricted path choices. When comparing RRT* with our approach, RRT* tends to pass through the center of areas densely populated with people, while our method actively avoids these regions, producing a more socially-aware path. This is supported by the graph in Fig.~\ref{fig:graph-simple}, which highlights that, despite a longer path, the cumulative social cost is consistently lower with our method compared to the RRT*.

The results demonstrate that our method effectively navigates the robot while minimizing proximity to people whenever possible. However, in narrow corridors, limited path options force the robot to move closer to individuals. Compared to the standard RRT*, which often cuts through densely populated areas, our approach actively avoids these regions, generating more socially-aware paths. This is further supported by Figure~\ref{fig:graph-simple}, which shows that, despite producing longer paths, all social methods consistently achieve lower social costs than the baseline RRT*.

\subsection{Quantitative results}
\label{sec:c}

\begin{figure}[htbp]
    \centering
    \begin{subfigure}{0.8\linewidth}
        \centering
        \includegraphics[width=\linewidth]{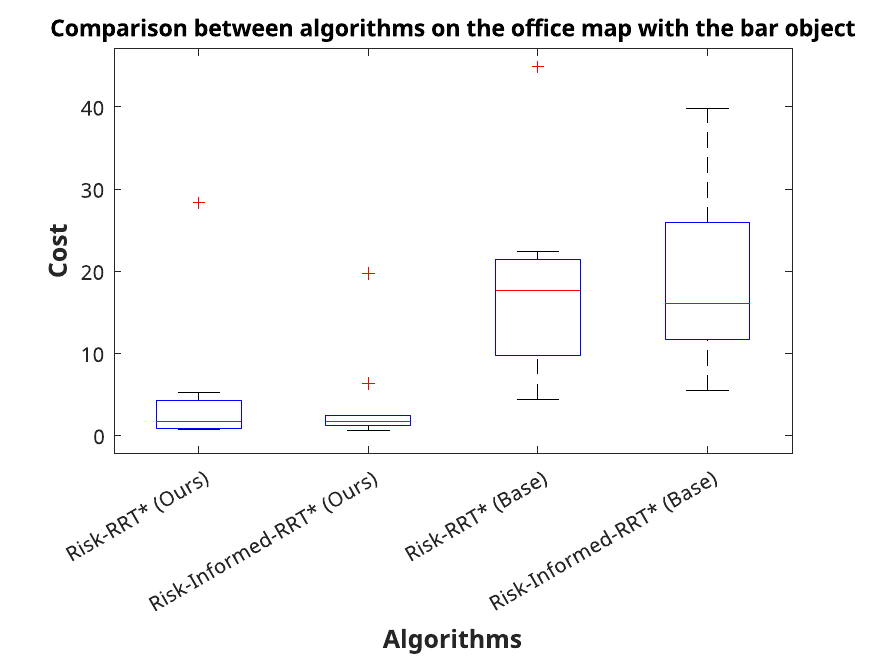}
        \caption{Office-like environment with Bar object.}
        \label{fig:office_boxplot}
    \end{subfigure}
    \begin{subfigure}{0.8\linewidth}
        \centering
        \includegraphics[width=\linewidth]{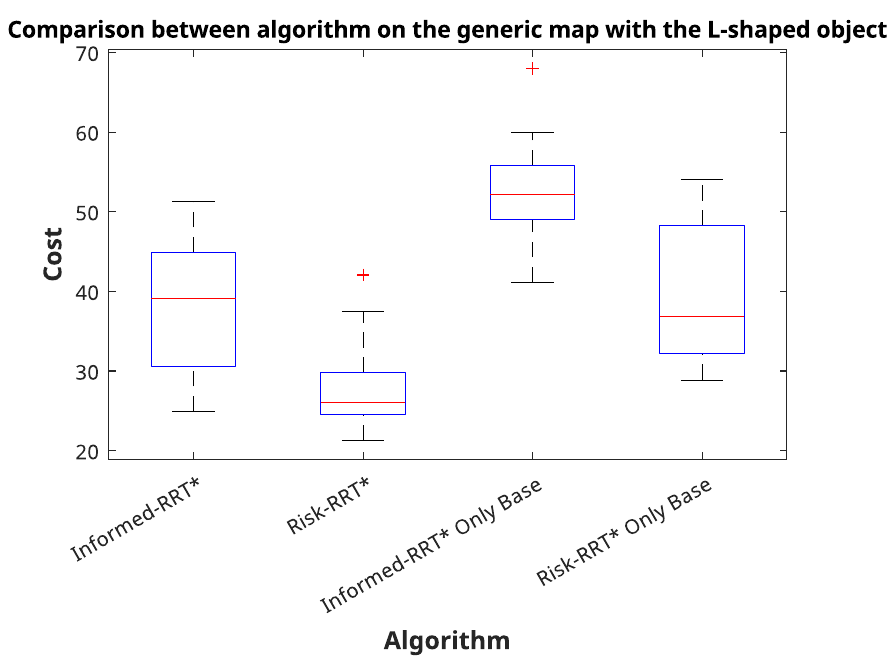}
        \caption{Generic environment with L-shaped object.}
        \label{fig:generic_boxplot}
    \end{subfigure}
    \caption{Boxplots of the final cost $\mathcal{F}(\mmconf{end})$ for each run of the social algorithms. The Y-axis represents the cost, while the X-axis indicates the algorithm applied.
    %\rev{Boxplots comparing the final costs for two scenarios, the y axis shows the final cost, and the x axis shows the algorithm used. The final costs were all calculated considering the robot base, the manipulator and the object, but the \textit{base} algorithms consider only the robot base for path plannig}
    }
    \label{fig:boxplot}
\end{figure}

%\rev{We quantitatively evaluate the four previously described social algorithms, comparing their performance both among themselves and, in particular, against their base-only counterparts. For such, we present and analyze boxplots of the 10 resulting path costs, i.e., $\mathcal{F}(\mmconf{goal})$.}
We quantitatively evaluate the four previously described social algorithms, comparing their performance against each other and their base-only counterparts. To this end, we analyze boxplots of the 10 resulting path costs, i.e., $\mathcal{F}(\mmconf{goal})$.
The results for the office-like environment (Section~\ref{sec:a}) are shown in Figure~\ref{fig:office_boxplot}, while those for the generic environment (Section~\ref{sec:b}) appear in Figure~\ref{fig:generic_boxplot}.

In the office scenario, the multi-point Risk-Informed-RRT* achieved the best performance, closely followed by the multi-point Risk-RRT*. However, in the generic environment, the multi-point Risk-RRT* significantly outperformed the informed variant.  

This behavior is expected: Informed approaches become highly biased toward the first path found since sampling is restricted after the initial solution. In contrast, the original Risk-RRT* continues exploring the entire space, even after identifying a feasible path. As a result, it has a higher chance of discovering longer but more socially acceptable paths. Once an informed variant is biased by a suboptimal initial path, it is less likely to sample and identify better alternatives.

Overall, it is clear that in both scenarios, the multi-point approaches consistently outperformed their base-only counterparts, demonstrating the superiority of our proposed method and the need for specific mobile manipulator techniques for socially-aware mobile manipulators.

% \begin{figure}[htpb]
%     \centering
%     \includesvg[width=.9\linewidth]{img/office_boxplot.svg}
%     \caption{Boxplot Office}
%     \label{fig:office_boxplot}
% \end{figure}

% \begin{figure}[htpb]
%     \centering
%     \includesvg[width=.9\linewidth]{img/generic_boxplot.svg}
%     \caption{Boxplot Generic}
%     \label{fig:generic_boxplot}
% \end{figure}

\subsection{Influence of social cost weight}
\label{sec:d}

\begin{figure}[htbp]
    \centering
    \begin{subfigure}{0.32\linewidth}
        \centering
        \includegraphics[width=\linewidth, trim={0.75cm 0cm 1.25cm 0.1cm}]{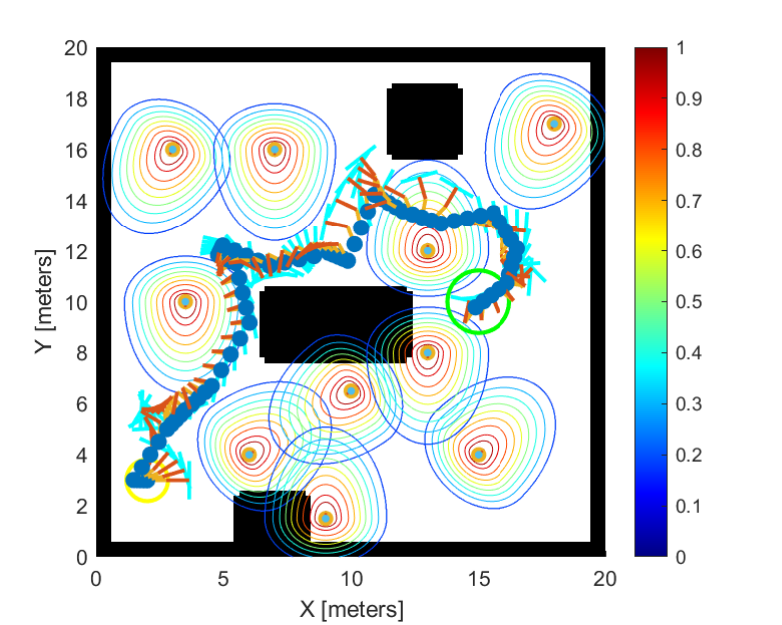}
        \caption{}
        \label{fig:W-object}
    \end{subfigure}
    \begin{subfigure}{0.32\linewidth}
        \centering
        \includegraphics[width=\linewidth, trim={0.75cm 0cm 1.25cm 0.1cm}]{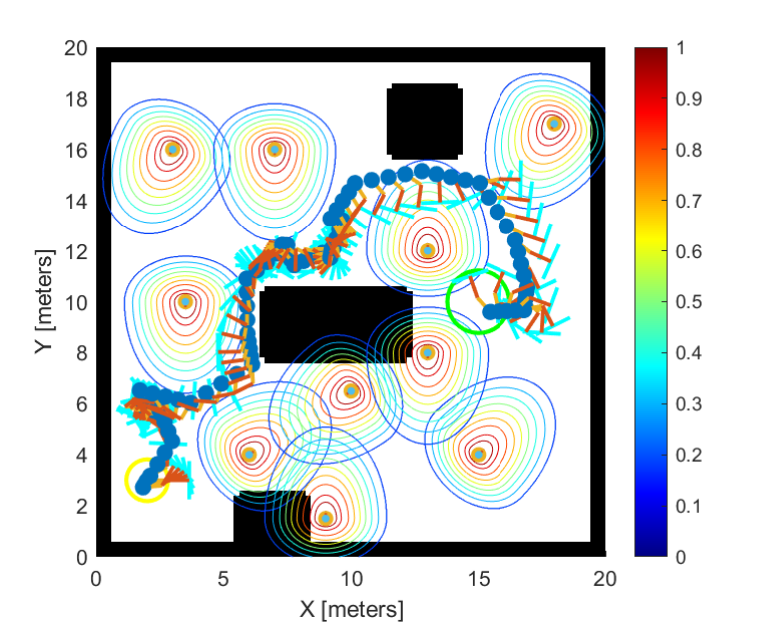}
        \caption{}
        \label{fig:W-base}
    \end{subfigure}
     \begin{subfigure}{0.32\linewidth}
         \centering
         \includegraphics[width=\linewidth, trim={0.75cm 0cm 1.25cm 0.1cm}]{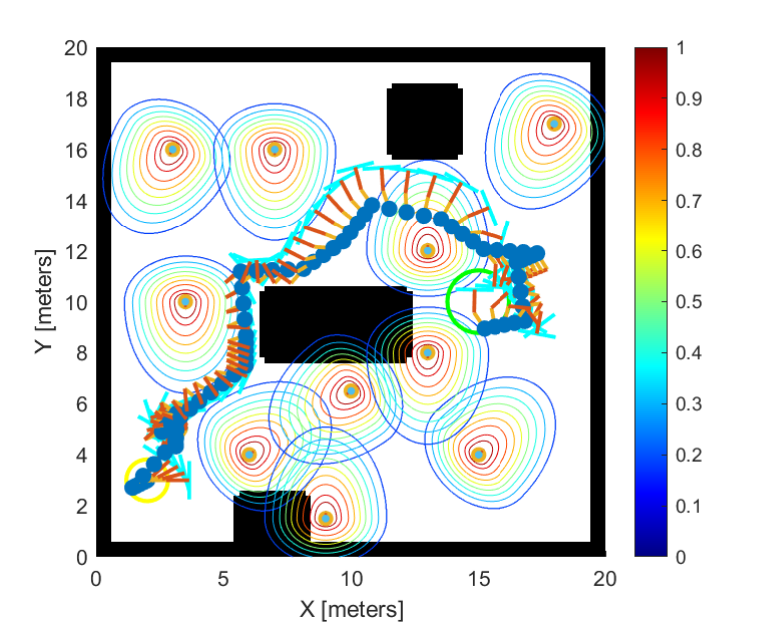}
         \caption{}
         \label{fig:W-equal}
     \end{subfigure}
    \caption{Social navigation results for different weights. (a) Weight of 0.95 for the object and 0.05 for the rest. (b) Weight of 0.95 for the base and 0.05 for the rest. (c) Equal weights.  Video: \href{https://youtu.be/tNo6pqw-BAA}{https://youtu.be/tNo6pqw-BAA}.}
    \label{fig:W-comparison}
    \vspace{-2mm}
\end{figure}

In this test, we used the generic map with the bar object. We evaluated three distinct weighting strategies. In the first case, we prioritized the points associated with the object, assigning them a weight of 0.95, while the remaining points were weighted at 0.05. In the second case, the focus shifted to the base, with the base point receiving a weight of 0.95 and the other points weighted at 0.05. Lastly, in the third scenario, we applied equal weights to all points, as in the previous sections. The best results are shown in Figure~\ref{fig:W-comparison}.

%Upon analyzing the paths, 
We observe that assigning greater weight to the base leads it to maintain a larger distance from people, while the object moves closer, and vice versa when the object is prioritized. These behaviors align with expectations, showcasing the method’s flexibility. However, assigning equal weights yields similar results to prioritizing the object, as the object has more interest points than other parts, giving it an inherent advantage. This effect could be mitigated by rebalacing the weigths.
%In summary, our approach allows for prioritizing which parts of the robot to be more cautious around people.

% \subsection{Simulated Results}
% The best result from each algorithm was simulated in CoppeliaSim to assess feasibility using the control scheme defined in Section IV-C. These results, available at: $<link>$, show that all approaches produced coherent and complete paths.

\section{Conclusion and Future Work}

As robots are increasingly integrated into human-populated environments, the field of social robotics has gained significant importance. The development of socially acceptable behaviors in robots enhances their acceptance, reduces potential risks, and enables more natural and effective collaboration between humans and robots in shared spaces.

In this paper, we tackle the problem of socially-aware object transportation using a mobile manipulator. By incorporating a social cost function a with multi-interest-point evaluation into the planning process, we ensure that the robot not only respects personal spaces but also accounts for the positions of the mobile base, the manipulator arm, and the transported object. %\rmv{This approach allows the mobile manipulator to carry out tasks efficiently and safely, while adhering to social norms, ensuring a higher degree of acceptance in human-populated environments.}
Our experiments highlight the need for developing techniques specifically for mobile manipulators, as our method outperforms socially-aware base-only approaches and offers the flexibility to prioritize certain parts of the robot to be more cautious.

In future work, we plan to extend our approach to address the transportation in 3D space, allowing the mobile manipulator to handle a wider range of tasks. We also aim to explore dynamic scenarios involving moving people, enhancing the robot's ability to adapt to changing environments. Finally, we intend to expand the experiments to beyond simulations, using real robots in controlled environments.

%In future work, we plan to extend our approach to address more complex types of transportation in 3D space, allowing the mobile manipulator to handle a wider range of tasks. We also aim to explore dynamic scenarios involving moving people and obstacles, enhancing the robot's ability to adapt to changing environments. Additionally, we intend to consider the arrangement of people in groups, which may require special considerations to ensure socially-acceptable behavior.

% \begin{itemize}
%     \item Dynamic people
%     \item 3D
%     \item Groups
%     \item Unexpected obstacles
% \end{itemize}
%%%%%%%%%%%%%%%%%%%%%%%%%%%%%%%%%%%%%%%%%%%%%%%%%%%%%%%%%%%%%%%%%%%%%%%%%%%%%%%%

%\section*{APPENDIX}

%\section*{Acknowledgment}

%%%%%%%%%%%%%%%%%%%%%%%%%%%%%%%%%%%%%%%%%%%%%%%%%%%%%%%%%%%%%%%%%%%%%%%%%%%%%%%%

%\clearpage
%\IEEEtriggeratref{11}

\bibliographystyle{IEEEtran}
\bibliography{bibliography}

\end{document}